\documentclass[acmsmall]{acmart}

\usepackage{booktabs}
\usepackage{rotating}
\usepackage{adjustbox}
\usepackage{diagbox}
\usepackage{multirow}
\usepackage{caption}
\usepackage[flushleft]{threeparttable} % tablenotes

\usepackage{graphicx}
\usepackage{epstopdf}
\graphicspath{{figures/}}
\usepackage{subfigure}
\usepackage{stfloats}

\usepackage{amsmath}
\usepackage{amsmath}
\usepackage{bm}
\usepackage{algorithm}
\usepackage{algorithmic}
\usepackage{listings}
\usepackage{array}

\def\BibTeX{{\rm B\kern-.05em{\sc i\kern-.025em b}\kern-.08emT\kern-.1667em\lower.7ex\hbox{E}\kern-.125emX}}

\begin{document}

%
% The "title" command has an optional parameter, allowing the author to define a "short title" to be used in page headers.
\title{Pose-aware Adversarial Domain Adaptation for Personalized \\ Facial Expression Recognition}

%
% The "author" command and its associated commands are used to define the authors and their affiliations.
% Of note is the shared affiliation of the first two authors, and the "authornote" and "authornotemark" commands
% used to denote shared contribution to the research.
\author{Shangfei Wang}\authornote{Dr. Shangfei Wang is the corresponding author.}
\email{sfwang@ustc.edu.cn}
\affiliation{%
  \institution{University of Science and Technology of China}
  \streetaddress{443 HuangShan Rd}
  \city{Hefei Shi}
  \state{Anhui Sheng}
  \country{China}
  \postcode{230027}
}
\author{Guang Liang}
\affiliation{%
  \institution{University of Science and Technology of China}
  \streetaddress{443 HuangShan Rd}
  \city{Hefei Shi}
  \state{Anhui Sheng}
  \country{China}
  \postcode{230027}
}
\email{xshmlgy@mail.ustc.edu.cn}

\author{Can Wang}
\affiliation{%
	\institution{University of Science and Technology of China}
	\streetaddress{443 HuangShan Rd}
	\city{Hefei Shi}
	\state{Anhui Sheng}
	\country{China}
	\postcode{230027}
}
\email{canwang@mail.ustc.edu.cn}

%
% By default, the full list of authors will be used in the page headers. Often, this list is too long, and will overlap
% other information printed in the page headers. This command allows the author to define a more concise list
% of authors' names for this purpose.
%\renewcommand{\shortauthors}{Trovato and Tobin, et al.}

%
% The abstract is a short summary of the work to be presented in the article.
\begin{abstract}
  Current facial expression recognition methods fail to simultaneously cope with pose and subject variations.
  In this paper, we propose a novel unsupervised adversarial domain adaptation method which can alleviate both variations at the same time. Specially, our method consists of three learning strategies: adversarial domain adaptation learning, cross adversarial feature learning, and reconstruction learning. The first aims to learn pose- and expression-related feature representations in the source domain and adapt both feature distributions to that of the target domain by imposing adversarial learning. By using personalized adversarial domain adaptation, this learning strategy can alleviate subject variations and exploit information from the source domain to help learning in the target domain.
  The second serves to perform feature disentanglement between pose- and expression-related feature representations by impulsing pose-related feature representations expression-undistinguished and the expression-related feature representations pose-undistinguished.
  The last can further boost feature learning by applying face image reconstructions so that the learned expression-related feature representations are more pose- and identity-robust.
  Experimental results on four benchmark datasets demonstrate the effectiveness of the proposed method.
\end{abstract}

%%
%% The code below is generated by the tool at http://dl.acm.org/ccs.cfm.
%% Please copy and paste the code instead of the example below.
%%
%</ccs2012>

%\end{CCSXML}

%\ccsdesc[500]{Computing methodologies~Computer vision}

%%
%% Keywords. The author(s) should pick words that accurately describe
%% the work being presented. Separate the keywords with commas.
%\keywords{Pose-aware, Adversarial Domain Adaptation, Facial Expression Recognition}

%%
%% This command processes the author and affiliation and title
%% information and builds the first part of the formatted document.
\maketitle

\section{Introduction}
Facial expression recognition has become a research hotspot. It has a wide range of applications, such as judgment of fatigue driving, criminal investigation and interrogation.
%, data collection and research related to emotion theory.
However, pose variation is often overlooked, that can badly influence the performance of facial expression recognition.
And there also exists subject variation between different subjects \cite{li2018deep}. This limits the application of facial expression recognition in real situations.
Therefore, researchers try to cope with pose and subject variations.

Apparently there are differences in the distribution of pose features and facial expression features \cite{li2019self}. In order to solve the problem of the influence of pose bias on facial expressions, researchers have proposed three categories of methods: single classifier method \cite{zhang2018joint}, pose normalization method \cite{lai2018emotion,jampour2017pose}, and pose-robust features method \cite{zhang2016deep,eleftheriadis2015discriminative,mao2016hierarchical,wu2017locality}.
Single classifier method aims to directly train a classifier using large amount of facial images with different poses and expressions. It depends on large amount of data to learn a robust classifier. However, these data are not always available in real application.
Pose normalization method tries to convert non-frontal facial images into frontal images and then performs facial expression recognition. However, the quality of produced frontal images can not be easily guaranteed.
Pose-robust features method aims to find a pose-robust expression-related feature representation.
Compared to the above two methods, this method can avoid image generation and address pose variation in feature level.
Therefore, we would like to design a pose-robust features method.

Identity variation is also a critical problem.
%we need to overcome.
%A solution is to search identity-irrelevant feature representations \cite{meng2017identity,yang2018facial} and the other method is to construct person-specific model for every subject intuitively \cite{chu2016selective,yang2018identity}. For the first method, researchers need to design a constraint depending on identity-related image pairs. However, such image pairs are hard to come by.
An intuitive solution is to train models for every person based on identity characteristics~\cite{wang2018personalized}. However, it needs a large number of images for each person, and these data are not easy to obtain in practical application. To address it, researchers tend to utilize domain adaption. Domain adaption extracts information from source domain(i.e. subjects in training dataset) to improve learning in target domain(i.e., novel subjects), and thus alleviating the problem of limited labeled data in the target domain. Domain adaption is divided into supervised and unsupervised domain adaption. The method where a small number of expression labeled data are available in the target domain is called supervised domain adaptation, and unsupervised domain adaption doesn't require any expression labels for novel subjects. However, current domain adaption methods ignore the pose bias existing in the source domain or the target domain, resulting in failure of the practical application.
So we are eager to perform domain adaptation for identity bias and simultaneously address pose bias.

Even above methods focus on either pose or subject variation. Few method takes both factors into account. In this paper, we propose a novel unsupervised domain adaptation method to simultaneously address both problems. As shown in Figure 1, the proposed method consists of three parts: adversarial domain adaptation learning, cross adversarial feature learning and reconstruction learning. The first contains a source branch and a target branch. The source encoder aims to extract pose- and expression-related feature representations via supervised training, respectively. The target encoder tends to extract pose- and expression-related feature representations by adversarial training with a pose domain discriminator and an expression domain discriminator, while the pose domain discriminator distinguishes which domain the pose feature belongs to and
so does the expression domain discriminator about the expression features.
Through adversarial learning, these domain discriminators aren't able to distinguish where these two features come from, so as to reduce the distribution difference between the source domain and the target domain without using any labeled target samples.
And the second is a cross discriminator which imposes adversarial training to perform feature disentanglement. And the feature disentanglement makes expression feature representation cannot be distinguished by pose discriminator and the pose feature representation cannot be distinguished by the expression classifier. Thus, these two feature representations can maintain their respective robustness. Last the generators is used to enhance constraints on feature representations by cross-combining pose and expression features from both domains.
By jointly training these learning strategies, our method achieves excellent results.

%The domain adaptation method is adopted to solve the lack of annotated data in the target domain while maintaining the subject invariance of extracted pose- and expression-related features. And the feature disentanglement keep the pose and make the pose feature away from the expression feature, so that the expression feature is not affected by it.
\section{Related Work}
\subsection{Pose-independent FER}
Pose independent facial expression recognition approaches fall into three categories: single classifier method, pose normalization method, and pose-robust features method.

Single classifier method aims to train a classifier using large amount of facial images with different poses and expressions. Zhang \emph{et al.} \cite{zhang2018joint} proposed a joint pose and expression modeling method (JPEM) to recognize facial expression. They use a generative adversarial network (GAN) to generate face images with different expressions under arbitrary poses. These generated images were then utilized to train the classifier. This method mainly depends on a large amount of generated images. However, it cannot capture relations between different poses. In addition, there is still room for improvement because of the difference between faked and real images. In contrast, our method can avoid using unrealistic faked images for training the classifier.

Pose normalization method tries to convert non-frontal facial representations into frontal representations and then performs facial expression recognition. Lai \emph{et al.} \cite{lai2018emotion} proposed a GAN-based multi-task learning method that learns emotion-preserving representations in the face frontalization framework. This method can transform a profile facial image into a frontal face via GAN while maintaining the same identity and expression. Then the expression classifier is trained using these generated images. However, due to the presence of the stretched artifact of the generated frontal face images, the recognition performance of large-angle face images is degraded. For another, Jampour \emph{et al.} \cite{jampour2017pose} proposed a pose mapping method for conversion of profile faces. This method focuses on the relationship between pose features, which is different from the previous method that addresses relationship in pixel-level. However, Jampour \emph{et al.}'s work needs image pairs for training, which are not easy to be obtained in real situations. Compared with pose normalization methods, the proposed method avoids generating faked images and doesn't rely on image pairs.

As for the pose-robust features method, Zhang \emph{et al.} \cite{zhang2016deep} proposed a deep neural network (DNN)-driven feature learning method. They extracted scale invariant feature transform (SIFT) features from each facial image. Then SIFT features are used as the input of DNN to imitate the neural cognition mechanism, and a projection layer was used to learn discriminative features across landmarks of different faces. Eleftheriadis \emph{et al.} \cite{eleftheriadis2015discriminative} proposed a discriminative shared Gaussian process latent variable model (DS-GPLVM) for multi-view classification. In this method, they first train a discriminitive manifold shared by multiple head poses of a same facial expression. Then facial expressions are classified in the manifold. Mao \emph{et al.} \cite{mao2016hierarchical} proposed a pose-based hierarchical Bayesian model for multi-pose facial expression recognition. This method combined local pixel features and global high-level information to learn an intermediate representation. And they dealt with the multi-pose problem by sharing a pool of features with various poses. In addition, Wu \emph{et al.} \cite{wu2017locality} proposed a locality-constrained linear coding based bi-layer (LLCBL) method which is used to construct a bag-of-features model. They utilized the first layer overall features extracted by overall BoF model to estimate pose. Similarly, pose-related features are obtained by using view-specific BoF model. By combining these two kinds of features, this method can address pose bias. For above methods, pose estimation is necessary before expression recognition. However, once pose predictions are wrong, the performance of expression recognition will be significantly impaired. While the proposed method utilizes high-level features adversarial learning to avoid this.

\subsection{Subject-independent FER}
There are two categories of subject-independent facial expression recognition methods: the person-specific method and the subject-robust feature method.

Person-specific method tries to reduce identity bias by learning
person-specific models. Chu \emph{et al.} \cite{chu2016selective} came up with a transductive learning method, named Selective Transfer Machine (STM). This method personalizes a generic classifier by optimizing classifier and weights of the relevant training samples.  However, this method is proposed on the basis of a hypothesis that the data distribution of the target domain can be represented by the data of the source domain. Yang \emph{et al.} \cite{yang2018identity} proposed an identity-adaptive training algorithm for facial expression recognition. In this method, multiple models are trained for each facial prototypic expression. Then a conditional GAN is adapted to generate images of the same subject with different expressions which are utilized to train person-specific classifier
to perform facial expression recognition for each subject.
Similarly, Wang \emph{et al.} \cite{wang2018personalized} proposed a generative adversarial recognition network (GARN) to generate a large amount of facial images
that are similar to the target domain but retain AU patterns of the source domain. This enriches and enlarges the
target dataset. However, for these GAN-based methods, unrealistic faked images would result in failure of training the classifier. While the proposed method does not rely on faked images for training the classifier.

For the subject-robust features method, Yang \emph{et al.} \cite{yang2018facial} proposed a De-expression Residue Learning (DeRL) method, which extracts information of the expressive component when performing de-expression learning procedure. By conditional GAN, they learned a generative model that is responsible for generating image with any expressions. The DeRL is based on the assumption that a facial expression is made up of an expressive component and a neutral component. However, this assumption doesn't always work in the wild.
In contrast, the proposed method does not depend on this hypothesis.
Meng \emph{et al.} \cite{meng2017identity} proposed the identity-aware convolutional neural network(IACNN) for facial expression recognition. In this method, an identity-sensitive contrastive loss is adopted to learn subject-related features which are utilized to recognize expression with expression features. Two identical CNN streams are trained  simultaneously by minimizing the prediction errors and maximizing the expression and subject distance. However, the images of the unseen subject are un-annotated, so the contrastive loss can not extract information from the target domain. Like Meng \emph{et al.}'s work, Liu \emph{et al.} \cite{liu2017adaptive} proposed an adaptive deep metric learning method (ADML). They proposed cluster loss function combined with the subject-aware hard-negative mining and online positive mining scheme for subject-invariant facial expression recognition which alleviates the difficulty of threshold validation and anchor selection. However, this method cannot be used in unsupervised learning domain adaptation. In contrast, the proposed method is not restricted to any hypothesis and labeled target data.

\subsection{Adversarial domain adaptation}

Recently, many adversarial domain adaptation methods have been proposed. The joint adaptation network(JAN) \cite{long2017deep} utilizes adversarial training strategy to maximize joint maximum mean discrepancy to distinguish the distributions of the source and target domain. However, JAN depends on labeled target data. The main idea of the domain-adversarial neural network(DANN) \cite{ganin2016domain} is to get feature representation which is predictive of the source sample labels by adversarial domain adaptation. However, the DANN needs a large amount of samples in target domain, while the proposed method does not. In the adversarial discriminative domain adaptation method(ADDA) \cite{tzeng2017adversarial}, the generator and discriminator perform domain adaption by playing against each other. Through the GAN-loss, they get an approximation of the distribution of the source domain and the target domain. However, ADDA is not an end-to-end framework and depends on complex training strategy, while the proposed method is an end-to-end method. The deep cocktail network(DCTN) \cite{xu2018deep} utilizes weighted combination of source distributions to generate target distribution by imposing adversarial learning between each pair of source domain feature and adversary between source and target domain. The DCTN requires a lot of effort to train large amounts of classifiers and discriminators, while the proposed only needs to train a single network.
Furthermore, these adversarial domain adaptation methods can only be used to
address subject bias, but fail to simultaneously address pose bias.
Therefore, it can demonstrate the superiority of the proposed method.

To the best our knowledge, only the identity- and pose- feature learning method(IPFR)~\cite{wang2019identity} considers both pose and subject bias. IPFR leverages adversarial learning to yield feature representations that are good for expression recognition but do not differentiate the pose or the subject. However, this method
applies the same adversarial feature learning strategy to address both variations, so this method has to rely on accurate subject annotations during training. But samples related to a subject are very limited. Besides, IPFR tries to remove pose information from the expression-related feature representation, while the proposed method aims to disentangle expression- and pose-related feature representations, thus pose information for an unseen sample can also be obtained for a real-word application.
Furthermore, IPFR depends on the face recognition ability of the subject discriminator, which itself is a hard task.
%Currently few method considers both pose and subject bias. The proposed method considers both factors simultaneously.
By approximating the distribution of features in the source domain and the target domain, and increasing the distance between pose and expression features in feature space, the proposed method learns a robust feature representation of expression which removes information about pose and subject bias. Thus the proposed method can certainly improve the performance.% of facial expression recognition.

\begin{figure*}[htb]
      \centering
      \includegraphics[width=1\textwidth]{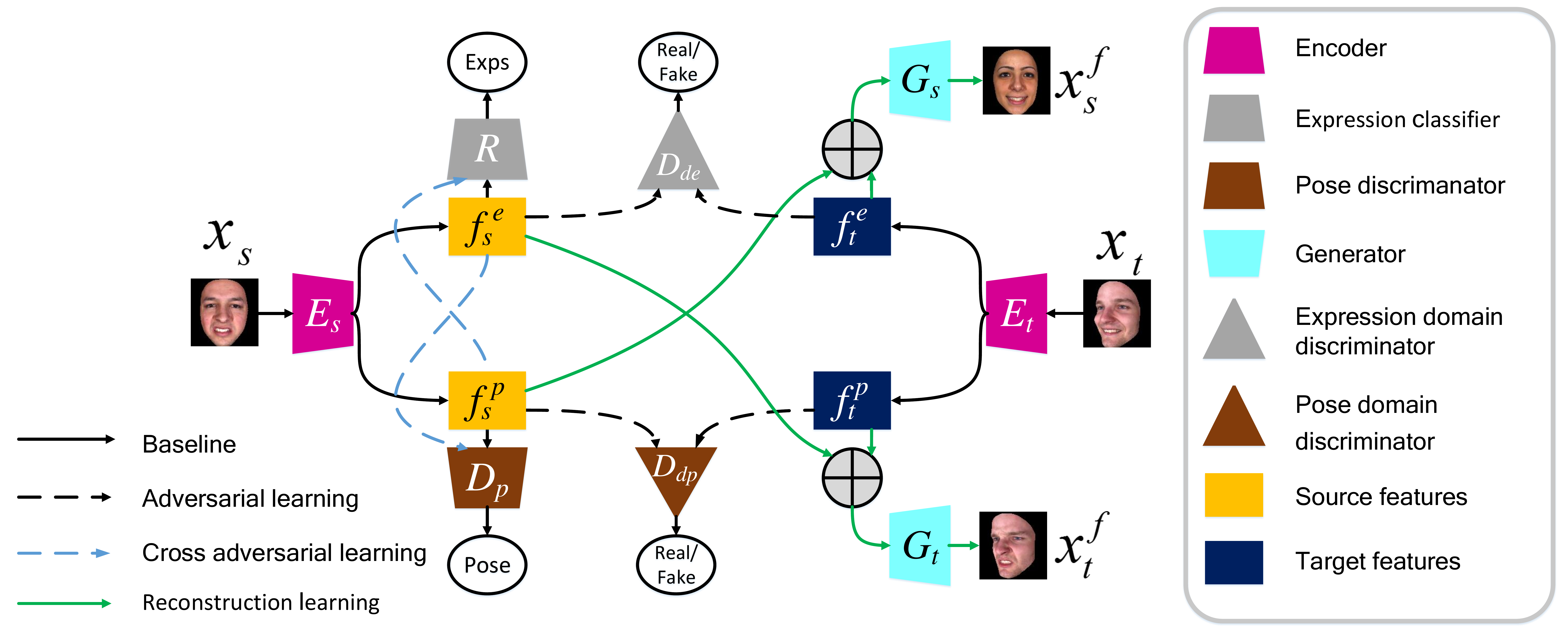}
      \caption{The structure of the proposed method. It contains eight components: a source  expression encoder $E_s$, a target encoder $E_t$, a source pose discriminator $D_p$, a source expression classifier $R$, an expression domain discriminator $D_{de}$, a pose domain discriminator $D_{dp}$, a source generator $G_s$, a target generator $G_t$.}
      \label{figurelabe1}
\end{figure*}

\section{Problem Statement}
Let $S=\left \{ \left ( x_s^i,y_s^i,p_s^i \right ) \right \}_{i=1}^{N}$ denote the samples from the source domain, where $x_s^i$ represents a source image, $y_s^i$ represents the expression label, $p_s^i$ represents the pose label, and $N$ is the number of source samples. $T=\left \{ \left ( x_t^i \right ) \right \}_{i=1}^{M}$ indicates the samples from the target domain, where $x_t^i$ represents a target image, and $M$ is the number of target samples. Source and target images are from different subjects. Given an unseen target image without any labels, the goal is to learn a classifier $R$ to predict its expression label.

Due to the limited data of target domain, i.e., $M\ll N$, we aim to exploit the source domain to help learning in the target domain. Meanwhile, pose variations can badly influence the domain adaptation and the expression recognition, thus we want to learn a pose-robust feature representation simultaneously.
Therefore, the goal is to minimize distance between the source and target mapping distributions, as show in Equation-\ref{Equ1}.
\begin{equation}
\label{Equ1}
\min Distance(p_s(f_s^e),p_t(f_t^e))
\end{equation}
where $f_s^e$ and $f_t^e$  represent the pose-robust expression-related feature representations of the source domain and the target domain, respectively.

\section{Methodology}
Figure-1 illustrates the framework. It contains eight components: a source encoder $E_s$, a target encoder $E_t$, a source pose discriminator $D_p$, a source expression classifier $R$, an expression domain discriminator $D_{de}$, a pose domain discriminator $D_{dp}$, a source generator $G_s$, and a target generator $G_t$.
In the adversarial domain learning, $E_s$ maps a source image to a pose-related feature representation $f_s^p$ and a expression-related feature representation $f_s^e$, where $f_s^e$ is pose-invariant. $D_p$ tries to predict the pose label of $f_s^p$. $E_t$ maps a target image to two features, $f_t^p$ and $f_t^e$, where $f_t^p$ is expected to share the same distribution to $f_s^p$ and $f_t^e$ is expected to share the same distribution to $f_s^e$.
$D_{de}$ tries to distinguish the domain of a feature representation $f_s^e$ or $f_t^e$, simultaneously $D_{dp}$ seeks to distinguish the domain of a feature representation $f_s^p$ or $f_t^p$, and $R$ tries to predict the expression label of $f_s^e$.
Meanwhile, in the cross adversarial learning, the $R$ and $D_p$ separates the distribution of the two features $f_s^p$ and $f_s^e$ by making $f_s^p$ undistinguished for $R$ and $f_s^e$ undistinguished for $D_p$.
In the reconstruction learning, given $f_s^p$ and $f_t^e$, $G_s$ tries to generate a faked source image $x_s^f$, while given $f_t^p$ and $f_s^e$, $G_t$ tries to generate a faked target image $x_t^f$.
Our goal is to learn pose-invariant feature representations $f_s^e$ and $f_t^e$ and minimize the distance between $p_s(f_s^e)$ and $p_t(f_t^e)$, thus a pose- and domain-robust expression classifier $R$ can be learned through $f_s^e$.

\subsection{Adversarial Domain Adaptation Learning}
The input of the framework is $(x_s,x_t,y_s,p_s)$ where $y_s$ represents expression labels of the source sample $x_s$, and the $p_s$ means the pose labels of the source sample $x_s$. The encoder take an image as input,
and outputs two kinds of feature representations, the pose- and expression-related feature representations.
To learn pose-related feature representations, pose discriminators are applied to $f_s^p$ and the loss can be defined as Equation-\ref{Equ2}.
\begin{equation}
\label{Equ2}
\ell_p(E_{s},D_{p})=\mathbb{E}\left [ \log(D_p(f_s^p)) \right ]
\end{equation}
Similarly, to learn expression-related feature representations, an expression classifier $R$ is applied to $f_s^e$. The expression recognition loss can be defined as Equation-\ref{Equ3}.
\begin{equation}
\label{Equ3}
\ell_e(E_{s},R)=\mathbb{E}\left [ \log(R(f_s^e)) \right ]
\end{equation}
Here a common classifier $R$ is expected to perform facial expression recognition on $f_s^e$ and $f_t^e$. However, due to the distribution difference between $p(f_s^e)$ and
$p(f_t^e)$, the expression contents involved in the source domain fail to help training the expression classifier $R$ for the target domain.
Thus, adversarial domain adaptation is introduced to minimize this distribution difference. Specifically, $E_s$, $E_t$ and $D_{de}$ play an adversarial game in which $E_s$ and $E_t$ try to minimize the divergence of feature distributions between $p_s(f_s^e)$ and $p_t(f_t^e)$ so that $D_{de}$ can not correctly recognize the domain of a sample. Similarly, $D_{dp}$ also play the adversarial game on $p_s(f_s^p)$ and $p_t(f_t^p)$ with $E_s$, $E_t$. The adversarial loss can be defined as Equation-\ref{Equ4}
\begin{equation}
\label{Equ4}
\begin{split}
\ell_{adv}(E_{s,t},D_{de,dp})=
\mathbb{E}\left [ \log(D_{de}(f_s^e)) \right ]+\mathbb{E}\left [ \log(1-D_{de}(f_t^e)) \right ]\\
+\mathbb{E}\left [ \log(D_{dp}(f_s^p)) \right ]+\mathbb{E}\left [ \log(1-D_{dp}(f_t^p)) \right ]
\end{split}
\end{equation}

By learning $E_{s,t}$, $D_{p}$, $D_{de}$, $D_{dp}$ and $R$ jointly, expression-related and domain-invariant feature representations $f_s^e$ and $f_t^e$ can be obtained for training a robust expression classifier $R$. The total goal is to optimize the following minmax objective, defined as Equation-\ref{Equ5}.
\begin{equation}
\label{Equ5}
\begin{split}
\min_{E_{s,t},D_{p},R}  \max_{D_{de,dp}} \ell_p(E_{s},D_{p})+\alpha \ell_e(E_{s},R)\\
-\beta \ell_{adv}(E_{s,t},D_{de,dp})
\end{split}
\end{equation}
where $\alpha$ and $\beta$ are weighted coefficients.

\subsection{Cross Adversarial Feature Learning}
To make sure $f_s^p$ preserves pose-related information without expression-related information and $f_s^e$ preserves expression information but avoids pose variations, we propose a cross adversarial feature learning method. Specifically, $E_s$ tries to generate $f_s^p$
that fail to be expression distinguished. Similarly, $E_s$ tries to generate $f_s^e$
that fail to be pose distinguished. Based on this intuition, we formulate the following objectives:
\begin{equation}
\label{Equ6}
\begin{split}
\max_{E_{s}}\ell_{cross}(E_{s},R,D_p) \\
\end{split}
\end{equation}
where the cross adversarial feature learning loss can be defined as Equation-\ref{Equ7}.
\begin{equation}
\label{Equ7}
\begin{split}
\ell_{cross}(E_{s},R,D_p)= \mathbb{E}\left [ \log(R(f_s^p)) \right ]+\mathbb{E}\left [ \log(D_p(f_s^e)) \right ]
\end{split}
\end{equation}
The goal of $E_s$ is to generate $f_s^p$ that are pose-related and expression-unrelated, and
$f_s^e$ that are expression-related and avoid pose variations. The proposed feature disentanglement technique allows more robust feature representations
for pose-robust facial expression recognition.

\subsection{Reconstruction Learning}
The idea of the reconstruction learning is to further improve the performance of the expression classifier by learning optimized feature representations $f_s^e$ and $f_t^e$.
If $f_s^e$ and $f_t^e$ can be expression-related and pose-unrelated,
after adding pose information, we can generate a corresponding expression- and pose-annotated facial image.
As shown in Figure-\ref{figurelabe1}, $f_t^e$ and $f_s^p$ are concatenated to reconstruct a source image $x_s^f$ and $f_s^e$ and $f_t^p$ are concatenated to reconstruct a target image $x_t^f$, defined as Equation-\ref{Equ8}
\begin{equation}
\label{Equ8}
\begin{split}
x_s^f=G_s(f_s^p,f_t^e) \\
x_t^f=G_t(f_t^p,f_s^e)
\end{split}
\end{equation}
where $x_s^f$ should be similar to $x_s^j\in S$, $x_s^j$ shares the same pose label as the input source sample $x_s$ and the same expression label as the input target sample $x_t$. $x_t^f$ should be similar to $x_t^k\in T$, where $x_t^k$ shares the same pose label as the input target sample $x_t$ and the same expression label as the input source sample $x_s$.
To sample $x_s^j$ from the source domain and $x_t^k$ from the target domain, pose and expression labels of $x_t$ should be used. To cope with it, we adopt pseudo labels for $x_t$, i.e., $\hat{y_t}=R(E_t(x_t))$ and $\hat{p_t}=D_p(E_t(x_t))$.
The training for the reconstruction learning is guided by an $\mathcal{L}_2$ reconstruction loss, which can be defined as Equation-\ref{Equ9}.
\begin{equation}
\label{Equ9}
\ell_{clc}(E_{s,t},G_{s,t})=\mathbb{E}\big [ \left \| x_s^f-x_s^j \right \|_2 \big ]+\mathbb{E}\big [ \big \|  x_t^f-x_t^k \big \|_2 \big ]
\end{equation}Note here we concatenate $f_s^p$ and $f_t^e$ rather than
$f_s^p$ and $f_s^e$ to generate $x_s^f$. This is because
$p(f_s^e)$ and $p(f_t^e)$ are expected to share the same distribution, and so do $p(f_s^p)$ and $p(f_t^p)$. And the same explanation for generating $x_t^f$.

\subsection{Overall Learning}
The joint objective function is defined as Equation-\ref{Equ10}.
\begin{equation}
\label{Equ10}
\begin{split}
\min_{E_{s,t},G_{s,t},R}   \max_{D_d,D_{de,dp}} \ell_p(E_{s},D_{p})+\alpha \ell_e(E_{s},R)\\
+\eta \ell_{clc}(E_{s,t},G_{s,t}) -\beta \ell_{adv}(E_{s,t},D_{de,dp})\\
-\gamma \ell_{cross}(E_{s,t},R,D_p)
\end{split}
\end{equation}
where $\alpha, \eta, \beta$ and $\gamma$ are weighted coefficients. Note only parameters of $E_s$ and $E_t$ are updated and others are fixed when optimizing
the loss $\ell_{cross}(E_{s,t},R,D_p)$,
The minmax optimization problem of Equation-\ref{Equ10} can be solved by applying an iterative algorithm, defined as Algorithm-\ref{alg:training}.
\begin{algorithm}[!htbp]
\footnotesize
\caption{The learning algorithm.}
\label{alg:training}
\begin{algorithmic}[1]
\REQUIRE The source and target training set $S$ and $T$; \\
         Training steps: $K_1$, $K_2$ and $K_3$; \\
         Batch size $m$ and the number of training epochs $K$.
\ENSURE The final classifier $f=E_t\circ R$.
\FOR {$t=1$ to $K$}
\FOR {$k=1$ to $K_1$}
\STATE Randomly sample $\left \{ x_s^i,y_s^i,p_s^i \right \}_{i=1}^{m}\sim S$.
\STATE Randomly sample $\left \{ x_t^i \right \}_{i=1}^{m}\sim T$.
\STATE Update $E_s$ and $R$ jointly: $\begin{aligned}\bigtriangledown  \theta_{E_{s}\cup R} := \frac{\partial \ell_e(E_{s},R)}{\partial \theta_{E_{s}\cup R}}\end{aligned}$.
\STATE Update $E_s$: $\begin{aligned}\bigtriangledown  \theta_{E_{s}} := - \frac{\partial \ell_{cross}(E_{s},R,D_p)}{\partial \theta_{E_{s}}}\end{aligned}$.
\STATE Update $E_s$ and $E_t$: $\begin{aligned}\bigtriangledown  \theta_{E_{s,t}} := -\frac{\partial \ell_{adv}(E_{s,t},D_{de,dp})}{\partial \theta_{E_{s,t}}}\end{aligned}$.
\ENDFOR
\FOR {$k=1$ to $K_2$}
%\STATE Randomly sample indicators $\left \{ z_i \right \}_{i=1}^{m}\sim \mathcal{Z}$.
\STATE Update $E_s$, $E_t$, $G_s$ and $G_t$ jointly: \\
$\begin{aligned}\bigtriangledown  \theta_{E_{s,t}\cup G_{s,t}} := \frac{\partial \ell_{clc}(E_{s,t},G_{s,t})}{\partial \theta_{E_{s,t}\cup G_{s,t}}}\end{aligned}$.
%\IF {$t>S$}
%\STATE Add faked images $\mathcal{D}^f=\left \{ x_i^f,y_i,z_i \right \}_{i=1}^{m}$ to $\mathcal{D}$:\\
%$\mathcal{D}=\mathcal{D} \cup \mathcal{D}^f$.
%\ENDIF
\ENDFOR
\FOR {$k=1$ to $K_3$}
%\STATE Randomly sample $\left \{ x_s^i,y_s^i,p_s^i \right \}_{i=1}^{m}\sim S$.
%\STATE Randomly sample $\left \{ x_t^i,y_t^i,p_t^i \right \}_{i=1}^{m}\sim T$, where $y_t^i=y_s^i$.
\STATE Update $D_p$ : $\begin{aligned}\bigtriangledown  \theta_{D_{p}} := \frac{\partial \ell_p(E_{s},D_{p})}{\partial \theta_{D_{p}}}\end{aligned}$.
\STATE Update $D_{de}$ and $D_{dp}$:\\
 $\begin{aligned}\bigtriangledown  \theta_{D_{de} \cup D_{dp}} := \frac{\partial \ell_s(E_{s,t},D_{de},D_{dp})}{\partial \theta_{D_{de} \cup D_{dp}}}\end{aligned}$.

\ENDFOR
\ENDFOR
\end{algorithmic}
\end{algorithm}

\section{Experiments}
Our experiments are conducted on four benchmark datasets: Multi-PIE \cite{gross2010multi}, BU-3DFE \cite{savran2008bosphorus}, AffectNet~\cite{mollahosseini2017affectnet} and SFEW~\cite{dhall2011static}.
\subsection{Experimental Conditions}
There are 755,370 images from 337 subjects which were taken under 15 viewpoints in the Multi-PIE database. The database contains six types of expressions: disgust, neutral, scream, smile, squint and surprise. Each facial image belongs to one of the six expressions.
We select 13,779 facial images of 100 subjects with all 9 poses~($\pm30^{\circ},\pm15^{\circ},0^{\circ},45^{\circ},60^{\circ},75^{\circ}$ and $90^{\circ}$ pan angles) following Wu \emph{et al.}~ \cite{wu2017locality} and Zhang \emph{et al.}~\cite{zhang2018joint}'s work.

\begin{table*}[]
\tabcolsep=6pt
\centering
\caption{Experimental results on the Multi-PIE and BU-3DFE databases.}
\label{table1}
\begin{footnotesize}
\scalebox{0.85}[0.80]{
\begin{tabular}{c|ccccccc|c|ccccc|c}
\hline
\multirow{2}{*}{Method} & \multicolumn{14}{c}{Multi-PIE}                                              \\ \cline{2-15}
                        & $0^{\circ}$ & $15^{\circ}$ & $30^{\circ}$ & $45^{\circ}$ & $60^{\circ}$ & $75^{\circ}$ & $90^{\circ}$ & Avg. & $-30^{\circ}$ & $-15^{\circ}$ & $0^{\circ}$ & $15^{\circ}$ & $30^{\circ}$ & Avg. \\ \hline
UPADA$_{R}$ & 88.6  &  89.7  &  84.4  &  79.5    &  77.2  &  76.7  &  72.7  &  81.3  & 82.7  &  87.0  &  88.5  &  87.5  &  81.6  &  85.5      \\ \hline
UPADA$_{R+adv}$ & 92.8  &  93.1  &  90.2  &  85.7  &  83.6  &  81.9  &  81.4  &  87.0  & 89.9  &  92.1  &  91.3  &  91.9  &  88.4  &  90.7      \\ \hline
UPADA$_{R+adv+cross}$        & 93.7  &  94.2  &  92.7  &  87.6  &  86.7  &  83.7  &  83.6  & 88.9   & 92.7  &  93.7  &  92.3  &  93.4  &  91.1  &  92.6   \\ \hline
UPADA       & 93.7  &  94.8  &  92.8  &  89.2  &  87.5  &  84.3  &  83.7  &\textbf{ 89.4}   & 92.7  &  94.1  &  93.8  &  93.9  &  92.4  &  \textbf{93.4}    \\ \hline
\multirow{2}{*}{Method}  & \multicolumn{14}{c}{BU-3DFE}                                                \\ \cline{2-15}
                        & $-45^{\circ}$ & $-30^{\circ}$ & $-15^{\circ}$ & $0^{\circ}$ & $15^{\circ}$ & $30^{\circ}$ & $45^{\circ}$ & Avg. & $0^{\circ}$ & $30^{\circ}$ & $45^{\circ}$ & $60^{\circ}$ & $90^{\circ}$ & Avg. \\ \hline
UPADA$_{R}$  & 73.3 & 77.1 & 80.9 & 78.9 & 79.7 & 80.7 & 72.8 & 77.6  &  78.1 & 77.5 & 72.6 & 69.7 & 68.4 & 73.3   \\ \hline
UPADA$_{R+adv}$ & 82.2 & 83.5 & 85.6 & 85.2 & 86.2 & 84.9 & 81.8 & 84.2   & 83.4 & 81.8 & 80.4 & 76.7 & 73.3 & 79.1   \\ \hline
UPADA$_{R+adv+cross}$        & 83.9 & 85.4 & 88.1 & 87.9 & 87.3 & 86.8 & 84.7 & 86.3   & 85.7 & 83.6 & 82.1 & 80.1 & 78.8 & 82.1    \\ \hline
UPADA        & 84.7 & 87.5 & 88.6 & 88.1 & 88.2 & 86.9 & 85.2 & \textbf{87.0}   & 86.3 & 84.1 & 83.5 & 81.9 & 79.7 & \textbf{83.1}  \\ \hline
\end{tabular}
}
\end{footnotesize}
\end{table*}

The BU-3DFE database is a synthetic database. There are 100 subjects, of which 56 subjects are females and 44 subjects are males. Each subject contains six facial expressions (anger, disgust, fear, happiness, sadness and surprise).
%Specially, these expressions vary in intensity because of the ways that are triggered differently.
In order to facilitate the comparison, we select images of all 100 subjects from $\pm45^{\circ},\pm30^{\circ},\pm15^{\circ},0^{\circ},60^{\circ}$ and $90^{\circ}$ pan angles and obtain 21600 facial images,
following Wu \emph{et al.}~ \cite{wu2017locality} and Zhang \emph{et al.}~\cite{zhang2018joint}'s work.
%The data of SFEW database is filtered from AFEW database by calculating key frames based on face point clustering. The most commonly used version in academia is SFEW 2.0. There are three parts in this database: Train(958 samples), Val (436 samples) and Test (372 samples). Each image is labeled to one of seven expression categories, i.e., anger, disgust, fear, neutral, happiness, sadness, or surprise.

%There are 420299 images annotated with 11 categories in the AffectNet database. Each image's label belongs to one of six expression categories (anger, disgust, fear, neutral, happiness, sadness and surprise) in this experiment. So we selected 283901 images. 300 images of them belongs to testing set and the rest made up of training sets. However the AffectNet database lacks pose annotations, so we use a ResNet trained by on the Multi-PIE and BU-3DFE database to label the pose of AffectNet's images with one of frontal or non-frontal views. This method is set up like the IPA2LT \cite{zeng2018facial}.
Following Wu \emph{et al.}~\cite{wu2017locality} and Zhang \emph{et al.}~\cite{zhang2018joint}'s work, we employ two experiment settings on the Multi-PIE and BU-3DFE databases:
$(0^{\circ},90^{\circ})$ and $(-30^{\circ},30^{\circ})$ pan angles on the Multi-PIE dataset and
$(0^{\circ},90^{\circ})$ and $(-45^{\circ},45^{\circ})$ pan angles on the BU-3DFE dataset.
The model is evaluated using leave-one-subject-out validation. It means for every experiment, the target domain is facial images of one subject, while the source domain comprises of facial images of the remain subjects from the same dataset.
In all experiments, facial images in target domain are split into two parts: one with 2/3 samples for training and the other with 1/3 samples for testing. Note no expression labels can be seen for the target domain.

Multi-PIE and BP4D datasets are collected in controlled environments, to further demonstrate the effectiveness of the proposed method under real conditions, we also conduct within-database experiments on a large in-the-wild dataset, AffectNet~\cite{mollahosseini2017affectnet} and cross-database experiments on a small in-the-wild dataset, SFEW~\cite{dhall2011static}. For the AffectNet database, images from six
categories (anger, disgust, fear, neutral, happiness, sadness, and surprise) are utilized and 283,901 images are obtained, where 3,500 images as the testing set. This experimental
setting is as the same as IPFR. Because AffectNet lacks pose annotations, we train a ResNet on the BU-3DFE and Multi-PIE databases to annotate AffectNet with one of five views $(-30^{\circ},30^{\circ})$. For the SFEW dataset, we conduct two experiments: train on the BU-DFE dataset and test on the SFEW dataset, and train on the AffectNet dataset and test on the SFEW dataset. We report five times average ACC results on the both datasets.

We did four ablation studies. The first experiment is named UPADA$_{R}$, the baseline network which only trains $E_s$ and $R$ and applies both into the target domain without domain and cross adversarial and reconstruction learning.
We train $E_s$, $E_t$, $R$, $D_{de}$ and $D_{dp}$ using adversarial domain adaptation learning in the second experiment, named UPADA$_{R+adv}$.
The third experiment trains the same networks as the second experiment did but applying cross adversarial feature learning to UPADA$_{R+adv}$ to train these networks, named UPADA$_{R+adv+cross}$. As for the last experiment, we train all networks with applying reconstruction learning to UPADA$_{R+adv+cross}$, named UPADA.

\subsection{Experimental Results and Analyses}
Experimental results of ablation studies are shown in Table 1. From this table,
we obtain the followings:

Firstly, the UPADA$_{R+adv}$ has higher accuracy than the baseline method UPADA$_{R}$
%which does not take into account the pose and subject bias.
%For example, UPADA$_{R+adv}$ is 5.1\% more accurate than the baseline UPADA$_{R}$ in the $(-45^{\circ},45^{\circ})$ pan angles on the Multi-PIE database, and
For example, UPADA$_{R+adv}$ is 6.6\% more accurate than UPADA$_{R}$ in the $(-45^{\circ},45^{\circ})$ pan angles on the BU-3DFE dataset.
It indicates that after utilizing adversarial domain adaptation to minimize the pose and expression distribution difference respectively between the source and the target domain, we can alleviate the domain shift, leading to better results.

Secondly, the method UPADA$_{R+adv+cross}$ gives better results than UPADA$_{R}$ and UPADA$_{R+adv}$.
For instance, \\
UPADA$_{R+adv+cross}$ outperforms UPADA$_{R}$ by 7.1\% and 7.6\% respectively on the Multi-PIE database with five and seven pan angles. Furthermore, UPADA$_{R+adv+cross}$ also shows improvements compared to UPADA$_{R+adv}$ on the Multi-PIE and the BU-3DFE dataset.
This demonstrates that combining
adversarial domain adaptation learning and cross adversarial learning can boost the feature disentanglement, resulting in better results.

We design a reconstruction learning to further boost the feature learning process.
For example, UPADA outperforms UPADA$_{R+adv+cross}$ by 0.8\% and 0.5\% respectively on the Multi-PIE dataset with five and seven pan angles.
%, and 1\% and 1\% respectively on the BU-3DFE dataset with five and seven pan angles.
These improvements suggest that the reconstruction learning can further boost the domain adaptation and the feature disentanglement. Therefore, the proposed method achieves best results compared to other ablation studies.

\begin{figure}[htb]
      \centering
      \includegraphics[width=0.48\textwidth]{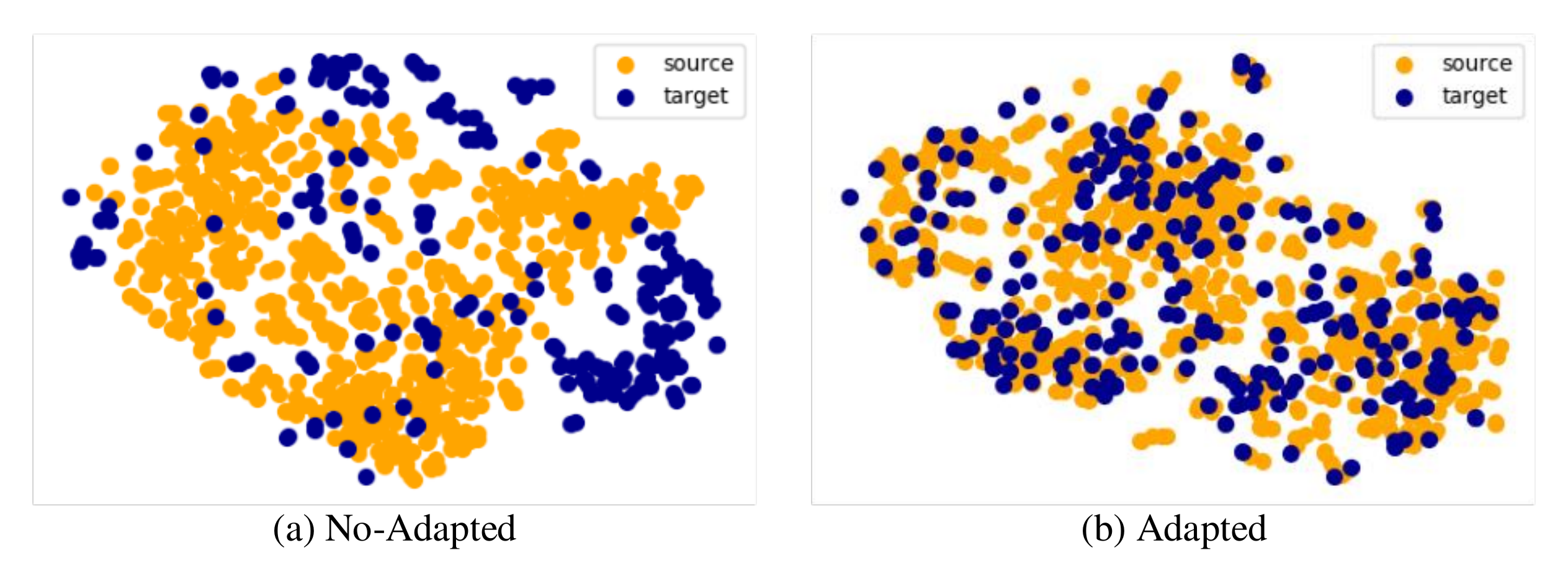}
      \caption{The effect of adaptation on the distribution of the extracted features. }
      \label{figurelabe2}
\end{figure}

\begin{figure}[htb]
      \centering
      \includegraphics[width=0.5\textwidth]{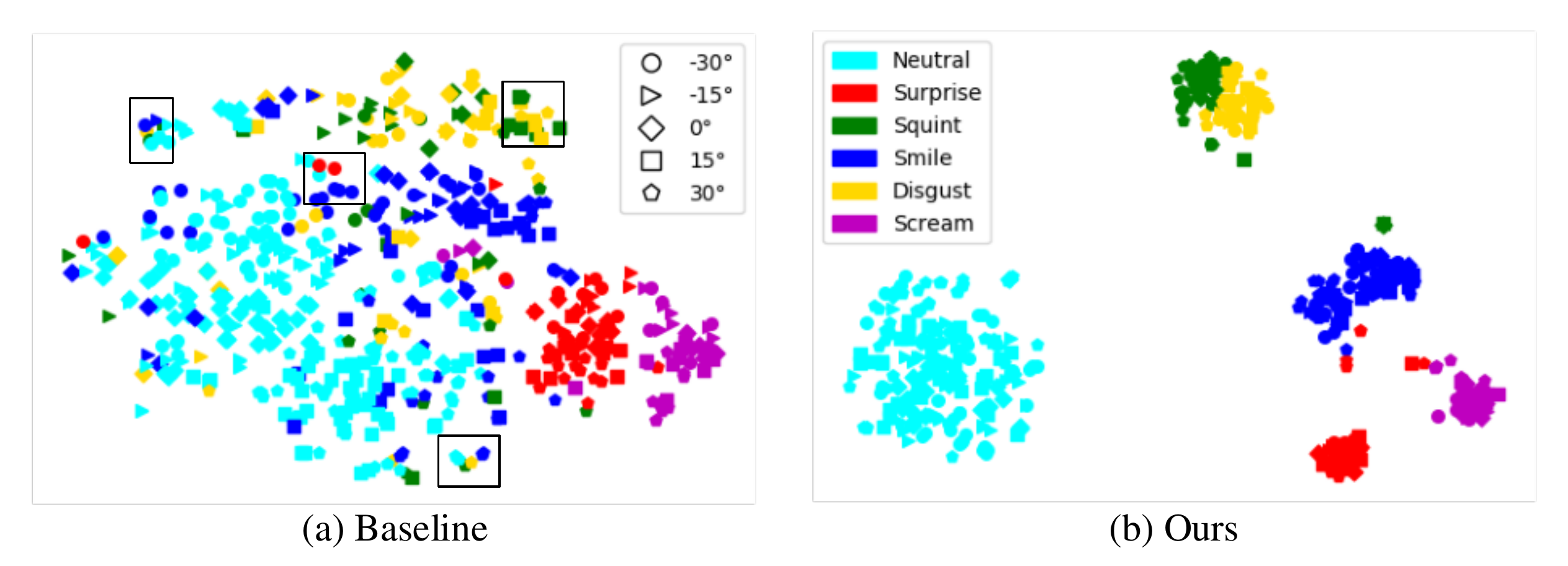}
      \caption{A visualization study. Different colors represent different expressions, while different shapes represent different poses.}
      \label{figurelabe3}
\end{figure}

\subsection{Analysis of Representation Learning}
In order to further demonstrate the effectiveness of our method for feature representation learning, we randomly select 2:1 source and target samples from the Multi-PIE dataset and use t-SNE to visualize the distribution of features extracted by the baseline method UPADA$_R$ and the proposed method UPADA, respectively, as shown in Figure-2.
Due to limited space, we do not show figures of the BU-3DFE dataset.
But similar observations can also be obtained of the BU-3DFE dataset.
From Figure-\ref{figurelabe2}-(a), there is clear
boundary between the source domain and the target domain of the baseline method without considering domain adaptation, suggesting that domain shift (subject bias) exists between both domains.
From Figure-\ref{figurelabe2}-(b), samples from the source and the target domain scatter randomly and can not be distinguished, indicating the proposed method can alleviate that domain shift between the source and the target domain.
From Figure-\ref{figurelabe3}-(a), features extracted from the baseline method UPADA$_R$ can not be easily expression-distinguished due to the situation that samples of the same pose but different expressions tend to converge into a cluster (e.g, four close-ups enclosed by the black rectangles). This shows the poor performance of the baseline because of pose bias.
While from Figure-\ref{figurelabe3}-(b), the features extracted by UPADA are clearly expression-distinguished. Samples
of different expressions merge into different clusters, while
samples of the same pose are randomly distributed within these clusters, indicating UPADA can alleviate pose variations and achieve better performance.

\begin{figure}[htb]
      \centering
      \includegraphics[width=0.48\textwidth]{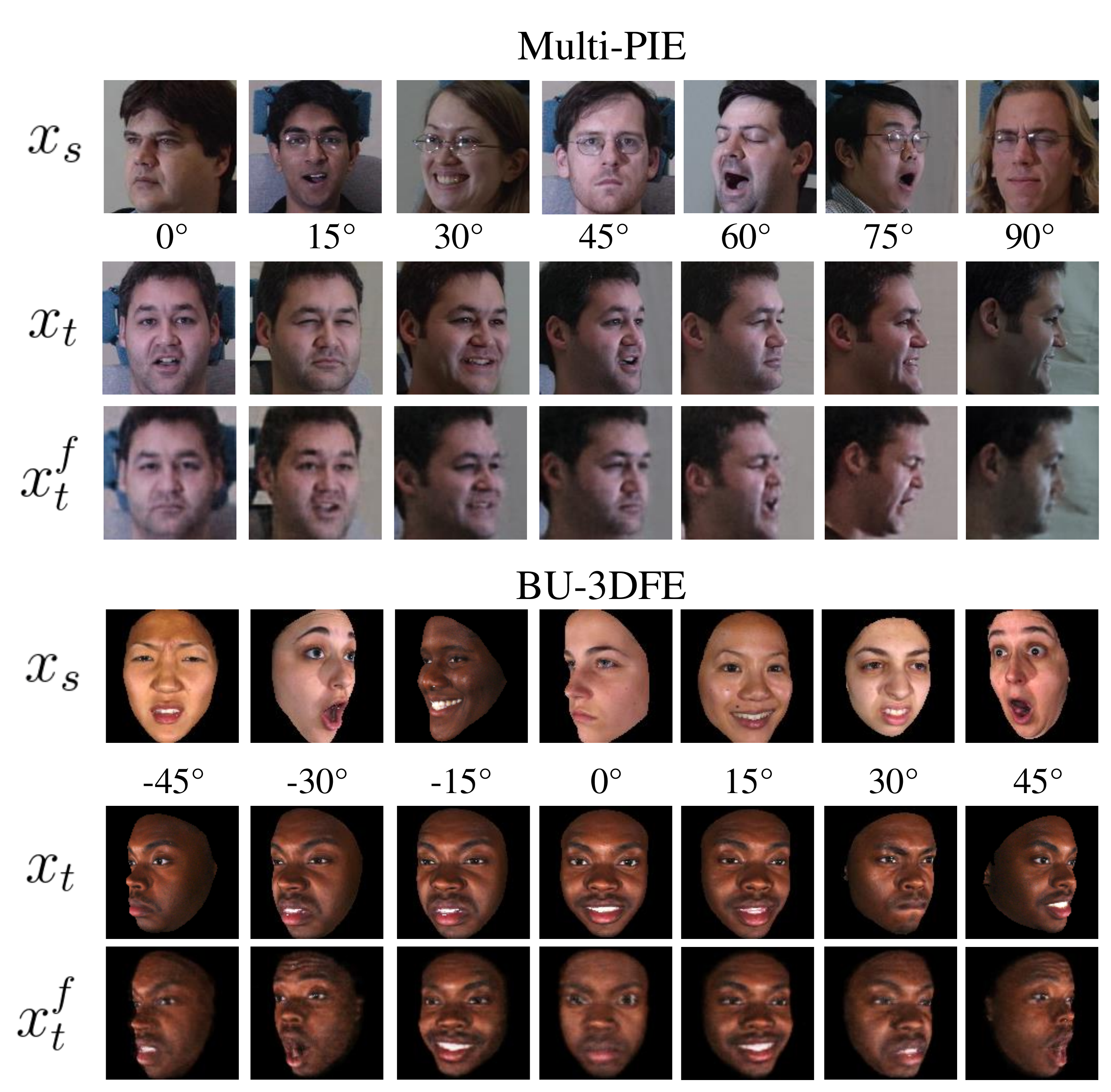}
      \caption{Generated images on the Multi-PIE and BU-3DFE databases.}
      \label{figure4}
\end{figure}

\subsection{Generator Analysis}
Reconstruction learning is designed to further boost the feature learning of adversarial domain adaptation and cross adversarial learning. Generators $G_s$ and $G_t$ are used to generate corresponding facial images.
Due to limited space, here we just show generated images of the target domain on the Multi-PIE and the BU-3DFE databases, as shown in Figure-\ref{figure4}.
Given the pose information of pose-related features $f_s^p$ from the source domain and expression information of expression-related features $f_t^e$ from the target domain, the generator $G_s$
is expected to generate
corresponding images and similar function for $G_t$.
It proves that the extracted feature representation successfully preserves the information related to each attribute.
Thus, the proposed reconstruction learning can perform well to
augment the performance of feature learning, leading to better expression recognition results.

\subsection{Comparison to Related Works}
%The methods involved in the comparison fall into three broad categories: pose-invariant methods, identity-invariant methods, and adversarial domain adaptation methods.
For pose-invariant methods, we compare the proposed method to DNND \cite{zhang2016deep}, DS-GPLVM \cite{eleftheriadis2015discriminative}, EPRL \cite{lai2018emotion}, KPSNM \cite{jampour2017pose}, HBTM \cite{mao2016hierarchical}, LLCBL \cite{wu2017locality}, IPFR~\cite{wang2019identity} and JPEM \cite{zhang2018joint}. For identity-invariant methods, we compared the proposed method to ADML \cite{liu2017adaptive}, DeRL \cite{yang2018facial}, GARN \cite{wang2018personalized}, IACNN \cite{meng2017identity}, IPFR~\cite{wang2019identity} and IA-gen \cite{yang2018identity}. For adversarial domain adaptation methods, we compared the proposed method to ADDA \cite{tzeng2017adversarial}, DANN \cite{ganin2016domain}, JAN \cite{long2017deep} and DCTN \cite{xu2018deep}. We also compare to IPA2L~\cite{zeng2018facial}. IPA2LT proposed an end-to-end trainable network LTNet embedded with a scheme of discovering the latent truth from multiple inconsistent labels and the input images. However, this method totally ignores the pose and subject bias.
Experimental results of the pose-invariant methods are directly copied
from the original papers. For identity-invariant methods, we implement the ADML, IACNN and GARN for comparison. While experimental results of others are also directly copied from the original papers.
For adversarial domain adaptation methods, we implement ADDA and DCTN for comparison and use the codes provided by the authors of DANN and JAN to evaluate the performance on the Multi-PIE and BU-3DFE dataset.
Experimental results are shown in Table 1, Table 2 and Table 3.

\begin{table}[]
\centering
\begin{footnotesize}
\caption{Comparison to state-of-the-art methods on the Multi-PIE and BU-3DFE databases.}
\label{table2}
\scalebox{0.80}[0.75]{
\begin{tabular}{c|c|ccc|ccc}
\hline
\multirow{2}{*}{Methods} & \multirow{2}{*}{Feature} & \multicolumn{3}{c|}{Multi-PIE} & \multicolumn{3}{c}{BU-3DFE} \\ \cline{3-8}
                         &                          & Poses    & Pose Pan    & Acc   & Poses   & Pose Pan   & Acc   \\ \hline
KPSNM    & HoG+LBP    & 7    & $(0^{\circ},90^{\circ})$   & 83.1  & 5       & $(0^{\circ},90^{\circ})$        & 78.8   \\ \hline
LLCBL    & Dense SIFT   & 7    & $(0^{\circ},90^{\circ})$   & 86.3  & 5       & $(0^{\circ},90^{\circ})$        & 74.6   \\ \hline
DNND    & SIFT   & 7    & $(0^{\circ},90^{\circ})$   & 85.2  & 5       & $(0^{\circ},90^{\circ})$        & 80.1   \\ \hline
EPRL    & DNN   & 7    & $(0^{\circ},90^{\circ})$   & 87.1  & 5       & $(0^{\circ},90^{\circ})$        & 73.1   \\ \hline
GARN     & DNN   & 7    & $(0^{\circ},90^{\circ})$   & 80.0  & 5       & $(0^{\circ},90^{\circ})$        & 76.8   \\ \hline
IA-gen  & DNN   &  -    &  -     & -  & 1       & $0^{\circ}$        & 76.8   \\ \hline
DeRL  & DNN   &  -    &  -     & -  & 1       & $0^{\circ}$        & 84.2   \\ \hline
ADML    & DNN   & 7    & $(0^{\circ},90^{\circ})$   &  77.7 & 5       & $(0^{\circ},90^{\circ})$        &  72.1  \\ \hline
IACNN    & DNN   & 7    & $(0^{\circ},90^{\circ})$   & 76.4  & 5       & $(0^{\circ},90^{\circ})$        &  73.2  \\ \hline
ADDA    & DNN   & 7    & $(0^{\circ},90^{\circ})$   & 78.6  & 5       & $(0^{\circ},90^{\circ})$        & 73.3   \\ \hline
DANN & DNN   & 7    & $(0^{\circ},90^{\circ})$   & 78.1  & 5       & $(0^{\circ},90^{\circ})$        & 72.5   \\ \hline
JAN & DNN   & 7    & $(0^{\circ},90^{\circ})$   & 79.3  & 5       & $(0^{\circ},90^{\circ})$        & 72.6   \\ \hline
DCTN & DNN   & 7    & $(0^{\circ},90^{\circ})$   & 88.0  & 5       & $(0^{\circ},90^{\circ})$        & 79.0   \\ \hline
IPFR & DNN   & 7    & $(0^{\circ},90^{\circ})$   & 88.4  & 5       & $(0^{\circ},90^{\circ})$        & 81.8   \\ \hline
Ours    & DNN   & 7    & $(0^{\circ},90^{\circ})$   & \textbf{89.4}  & 5       & $(0^{\circ},90^{\circ})$        & \textbf{83.1}   \\ \hline
\hline
DS-GPLVM    & Landmark   & 5    & $(-30^{\circ},30^{\circ})$   & 90.6  & -       & -        & -   \\ \hline
HBTM    & SIFT   & 5    & $(-30^{\circ},30^{\circ})$   & 90.2  & 7       & $(-45^{\circ},45^{\circ})$        & 79.1   \\ \hline
JPEM    & DNN   & 5    & $(-30^{\circ},30^{\circ})$   & 91.8   & 7       & $(-45^{\circ},45^{\circ})$        &  81.2   \\ \hline
ADML    & DNN   & 5    & $(-30^{\circ},30^{\circ})$   & 87.8  & 7       & $(-45^{\circ},45^{\circ})$        &   80.4 \\ \hline
IACNN    & DNN   & 5    & $(-30^{\circ},30^{\circ})$   & 85.5 & 7       & $(-45^{\circ},45^{\circ})$        &  77.2 \\ \hline
GARN     & DNN   & 5    & $(-30^{\circ},30^{\circ})$   & 83.0  & 7       & $(-45^{\circ},45^{\circ})$        & 78.3   \\ \hline
ADDA    & DNN   & 5    & $(-30^{\circ},30^{\circ})$   & 84.7  & 7       & $(-45^{\circ},45^{\circ})$        & 78.8   \\ \hline
DANN    & DNN       & 5    & $(-30^{\circ},30^{\circ})$   & 85.1  & 7       & $(-45^{\circ},45^{\circ})$        & 81.8   \\ \hline
JAN     & DNN        & 5    & $(-30^{\circ},30^{\circ})$   & 88.8  & 7       & $(-45^{\circ},45^{\circ})$        & 82.3   \\ \hline
DCTN    & DNN        & 5    & $(-30^{\circ},30^{\circ})$   & 89.7  & 7       & $(-45^{\circ},45^{\circ})$        & 82.7   \\ \hline
IPFR    & DNN   & 5    & $(-30^{\circ},30^{\circ})$   & 92.6  & 7       & $(-45^{\circ},45^{\circ})$        & 85.1  \\ \hline
Ours    & DNN   & 5    & $(-30^{\circ},30^{\circ})$   & \textbf{93.4}  & 7       & $(-45^{\circ},45^{\circ})$        & \textbf{87.0}   \\ \hline
\end{tabular}}
\end{footnotesize}
\end{table}

For single classifier method, the proposed UPADA outperforms JPEM.
%For example, the proposed method outperforms JPEM by 8.9\% on the BU-3DFE dataset with  seven pan angles.
JPEM depends on GANs to generate a large amount of facial images with different expressions
under arbitrary poses. Unrealistic generated facial images would add noise to train the classifier, resulting in worse results. The proposed can avoid this generation process by addressing pose variations at feature level, leading to better results.
Because of learning the optimized features and avoiding stretching artifacts, the proposed method is also more accurate than pose normalization methods(KPSNM and EPRL).
%Specifically, compared to KPSNM, the proposed method achieves 5.7\% improvements respectively on the Multi-PIE dataset with seven pan angles.
%Compared to EPRL, the proposed method achieves 1.7\% improvements on the Multi-PIE dataset with seven pan angles.
The proposed method also achieves better results compared to pose-robust features methods, i.e., LLCBL, DNND, DS-GPLVM and HBTM.
%For example, on the BU-3DFE dataset with five pan angles, the proposed method
%improves the perfromance from 74.6\% to 82.9\% comapred to LLCBL.
All methods except DNND depend on pose estimation, thus errors
in pose estimation then propagate to expression recognition. DNND depends on SIFT descriptions, which fails to perform well when the angle is greater than
$30^{\circ}$\cite{wu2013comparative}. While our method does not have these shortcomings, improving
the performance of recognition.

The proposed method performs better than the person-specifc methods (IA-gen and GARN). These GAN-based
methods are overly rely on generated images for classifier training; unrealistic images add noise to the classifier,
resulting in poor performance.
In contrast, the proposed method can alleviate pose and identity bias at feature level, without depending on faked images.
Furthermore, our achieves better performance than identity-robust features methods, i.e., IACNN, ADML, and DeRL.
The contrastive loss of IACNN suffers
from drastic data expansion when constructing image pairs from the training set. The triplet loss of ADML does not perform well in the case of limited identities. DeRL assumes that a face expression consists of both a neutral component
and an expressive component, which may not hold true in real-life situations. Weaknesses of these methods account for their inferior results compared to our method.
%in comparison to the proposed method.

To further evaluate the performance of the proposed
method, we compare to adversarial domain adaptation methods ADDA, JAN, DANN and DCTN. These domain adaptation methods can only address identity bias and fail to address pose variations, thus these methods can not perform well when facing to multi-pose facial expression recognition. In contrast, the proposed can simultaneously address both variations, leading to better performance.

We find the prosed method shows superior results to other methods except IPFR on the SFEW dataset while training on the AffectNet dataset. But experimental results on different expressions of IPFR are very imbalanced, while that of the proposed method are much better. This may because IPFR is very data-sensitive and when seeing limited data of one class, the bad result will appear. In contrast, the proposed method can benefit from the knowledge transfer ability of the domain adaptation and alleviate this problem.

\begin{table}[]
\centering
\begin{footnotesize}
\caption{Comparison with state-of-the-art methods on the in-the-wild databases.}
\label{table3}
\scalebox{0.78}[0.92]{
\begin{tabular}{c|c|c|ccccccc|c}
\hline
\multirow{2}{*}{Methods}& \multirow{2}{*}{Train} & \multirow{2}{*}{Test} & \multicolumn{7}{c|}{Emotions}  & \multirow{2}{*}{Avg.} \\ \cline{4-10}
             &       &     & AN & DI & FE & HA & NE & SA & SU &                       \\ \hline
DS-GPLVM     & BU-3DFE   &  SFEW & 25.9  &  28.2  &  17.2  &  43.0  &  14.0  &  33.3  &  11.0  &  24.7                    \\ \hline
JPEM     &  BU-3DFE   & SFEW & 30.9  &  22.0  &  19.6  &  50.9  &  19.2  &  28.0  &  15.5  &  26.6                     \\ \hline
GARN     &  BU-3DFE   & SFEW    &  25.9  &  23.8  &  16.2  &  38.1  &  12.6  &  26.8  &  12.0 &  22.2  \\ \hline
IPFR & BU-3DFE  & SFEW  & 27.3 & 28.9 & 24.3 & 38.7 & 19.7 & 26.2 & 31.4  &   28.1               \\ \hline
UPADA$_{R}$ & BU-3DFE &  SFEW   & 24.6  &  23.7  &  18.9  &  37.8  & 14.3  &  25.3  &  12.6  &  22.5                 \\ \hline
UPADA$_{R+adv}$ & BU-3DFE &  SFEW   & 31.2  &  29.9  &  23.1  &  43.1  & 20.1  &  30.8  &  28.8  &  29.6                  \\ \hline
UPADA$_{R+adv+cross}$ & BU-3DFE &  SFEW   & 32.6  &  31.2  &  24.1  &  44.2  & 21.5  &  33.2  &  29.4  &  30.9                 \\ \hline
Ours & BU-3DFE &  SFEW   & 33.1  &  31.2  &  24.8  &  45.3  & 21.7  &  33.4  &  29.8  &  \textbf{31.3}    \\
\hline
\hline
IPA2LT &  AffectNet   & SFEW  & -  &  -  &  -  &  -  & -  &  -  &  -  &   55.6                  \\ \hline
IPFR  & AffectNet &  SFEW   & 68.3  &  28.7  &  32.0  &  90.2  & 58.6  &  53.3  &  68.8  &   57.1                  \\ \hline
UPADA$_{R}$ & AffectNet &  SFEW   & 55.7  &  28.9  &  31.7  &  75.8  & 43.5  &  50.2  &  56.7  &  48.9                 \\ \hline
UPADA$_{R+adv}$ & AffectNet &  SFEW   & 57.9  &  31.2  &  39.8  &  80.2  & 52.7  &  52.4  &  61.1  &  53.6                  \\ \hline
UPADA$_{R+adv+cross}$ & AffectNet &  SFEW   & 62.1  &  35.9  &  40.3  &  81.3  & 54.2  &  52.8  &  67.2 &  56.3                  \\ \hline
Ours & AffectNet &  SFEW   & 64.2  &  37.1  &  40.6  &  84.8  & 55.1  &  53.0  &  68.3  &  \textbf{57.6}    \\
\hline
\hline
IPA2LT &  AffectNet   & AffectNet  & -  &  -  &  -  &  -  & -  &  -  &  -  &   56.5                  \\ \hline
IPFR  & AffectNet &  AffectNet   & 68.3  &  38.5  &  37.8  &  80.6  & 58.4  &  55.0  &  63.0  &  57.4                  \\ \hline
UPADA$_{R}$ & AffectNet &  AffectNet   & 56.5  &  29.7  &  29.3  &  76.8  & 49.8  &  53.3  &  59.8  &  50.7                  \\
\hline
UPADA$_{R+adv}$ & AffectNet &  AffectNet   & 62.8  &  36.5  &  35.7  &  79.2  & 55.4  &  57.1  &  63.3  &  55.7                  \\ \hline
UPADA$_{R+adv+cross}$ & AffectNet &  AffectNet   & 63.1  &  38.9  &  40.2  &  81.8  & 57.7  &  62.8  &  63.9  &  58.3                  \\ \hline
Ours & AffectNet &  AffectNet   & 63.1  &  41.2  &  42.1  &  82.0  & 57.8  &  62.8 &  64.3  &  \textbf{59.0}    \\
\hline
\end{tabular}}
\end{footnotesize}
\end{table}

%%%%%%%%%%%%%%%%%%%%%%%%%%%%%%%%%%%%%%%%%%%%%%%%%%%%%%%%%%%%%%%%%%%%%%%%%%%%%%%%
\section{Conclusion}
In this paper, we propose an unsupervised pose aware domain adaptation method
to simultaneously alleviate pose and subject bias. By adapt the pose and expression distribution between the source and target domain using adversarial learning and design a cross adversarial learning strategy, the proposed method can perform feature disentanglement and learn pose- and identity-robust features for facial expression recognition. By designing the reconstruction learning process, feature learning in the adversarial domain adaption learning and the cross adversarial learning can be further augmented. Better experiments prove the superiority of the proposed method over other state-of-the-art methods.

%%%%%%%%%%%%%%%%%%%%%%%%%%%%%%%%%%%%%%%%%%%%%%%%%%%%%%%%%%%%%%%%%%%%%%%%%%%%%%%%

%
% The acknowledgments section is defined using the "acks" environment (and NOT an unnumbered section). This ensures
% the proper identification of the section in the article metadata, and the consistent spelling of the heading.

%
% The next two lines define the bibliography style to be used, and the bibliography file.
\bibliographystyle{ACM-Reference-Format}
\bibliography{arxiv}

\end{document}